\documentclass[twoside,11pt]{article}

\usepackage{jmlr2e}
\usepackage{hyphenat}
\usepackage{float}
\usepackage{placeins}
\usepackage{color}
\usepackage[utf8]{inputenc}

\ShortHeadings{A Volumetric Convolutional Neural Network for Brain Tumor Segmentation}{Sherman}
\firstpageno{1}

\begin{document}

\title{A Volumetric Convolutional Neural Network for Brain Tumor Segmentation}

\author{\name Ryan Sherman \\
\addr Independent Deep Learning Researcher}


\maketitle

\begin{abstract}
Brain cancer can be very fatal, but chances of survival increase through early detection and treatment. Doctors use Magnetic Resonance Imaging (MRI) to detect and locate tumors in the brain, and very carefully analyze scans to segment brain tumors. Manual segmentation is time consuming and tiring for doctors, and it can be difficult for them to notice extremely small abnormalities. Automated segmentations performed by computers offer quicker diagnoses, the ability to notice small details, and more accurate segmentations. Advances in deep learning and computer hardware have allowed for high-performing automated segmentation approaches. However, several problems persist in practice: increased training time, class imbalance, and low performance. In this paper, I propose applying V-Net, a volumetric, fully convolutional neural network, to segment brain tumors in MRI scans from the BraTS Challenges. With this approach, I achieve a whole tumor dice score of 0.89 and train the network in a short time while addressing class imbalance with the use of a dice loss layer. Then, I propose applying an existing technique to improve automated segmentation performance in practice. 
\end{abstract}


\section{Introduction}

\subsection{Brain Cancer}

Brain cancer is a relatively common cancer with around 80,000 new cases each year in the United States alone \citep{nih}. The cancer can originate in the brain, or---in many cases---cancer from other areas of the body, such as the neck or spine spreads to the brain. Brain tumors are either classified as malignant or benign and are placed into four different categories---I, II, III, and IV--- which indicate the fatality of the tumor in ascending order. While not all of these tumors are malignant, benign tumors can still possess malignant qualities and act like a malignant tumor, and they also have the potential to turn malignant. The more deadly tumors leave patients with a five year survival rate of 5.5\% \citep{nbts}. Brain cancer may not be always be diagnosed once the tumor begins to progress, so treating and understanding the tumor is critical. Common therapies include surgery, chemotherapy, and radiotherapy. It is difficult for doctors to very clearly understand where tumors are located, where they are growing, and what treatment may be needed. In fact, more aggressive tumors are much harder to understand than lower-grade tumors. In addition to being fatal, brain cancer is initially the most costly cancer for a patient, with an average annual cost of \$150,000 \citep{nbts}. By conducting a thorough analysis of brain tumors, doctors can recommend the best steps for patients, which helps save money and lives.

\subsection{Medical Imaging for Brain Cancer}

Medical imaging technology has undergone significant advancement over the last few decades to provide doctors with clearer images of the patient's tissue, bone, etc. Magnetic Resonance Imaging (MRI), Computed Topography (CT), and Magnetic Resonance Spectroscopy (MRS) provide 3D reconstructions for doctors. Advancements which improved the accuracy of MRI imaging have caused its surge in popularity compared to other imaging techniques, and MRI's low emittance of radiation has led to its use for helping doctors understand and diagnose brain tumors \citep{pmid23961024}. Also, different types of MRI scans allow the focus on specific regions of the brain to provide doctors with detailed images of the tumor. Before surgeons can begin, a series of MRI scans of the brain must be taken to identify the exact size and location of the tumor in the brain. Once these scans are loaded onto a computer, doctors are responsible for manually segmenting tumors, or they may use software to perform an automated segmentation.

\subsection{Deep Learning for Brain Tumor Segmentation}

Over the last decade, advances in the deep learning field have been accompanied by the release of many public datasets --- enabling machine learning applications in everything from language translation to object recognition to protein folding prediction. Within these advancements, a significant portion of the field's research focuses on applying convolutional neural networks (CNNs) to various image-related tasks. CNNs have proven to be quite accurate at image classification, object detection, and image segmentation, and as convolutional neural networks continue to improve, the medical community will make greater use of the technology to diagnose and detect diseases through imaging \citep{DBLP:journals/corr/HeZRS15,Szegedy:2013:DNN:2999792.2999897,DBLP:journals/corr/RonnebergerFB15, 7298965}. 

After Stanford's high-performing pneumonia detection with deep convolutional neural networks \citep{DBLP:journals/corr/abs-1711-05225}, it appeared that deep learning is fit for medical imaging-related tasks. Particularly because any mistakes in a brain tumor segmentation can make surgery even more prone to error, and because of the frequency and volume of MRI scans taken for a patient, this is an ideal problem to be tackled by deep learning. 
Further: 

\begin{itemize}
  \item Doctors are prone to fatigue after looking at so many scans of their patients.
  \item They face time constraints which limit their ability to take their time and consider all the small details.
  \item The tumor shape, size, and modality can be vastly different from patient to patient, so patterns are more difficult to uncover for doctors.
\end{itemize}

For several years, computer vision experts have been developing and fine-tuning algorithms for brain tumor segmentation. Fairly recently, advances in deep learning and computing hardware have allowed for high precision in automated medical image segmentation.

Since 1995, many different approaches have been taken to automatically segment medical images, each with varying success. \cite{DBLP:journals/corr/abs-1710-04043} incorporated CNNs into a bounding box and scribble-based segmentation pipeline which produced fairly strong results. After performing a binary segmentation, a scribble-based segmentation refined the prior segmentation using either unsupervised or supervised learning. Using random forests, \cite{DBLP:journals/corr/SoltaninejadZLA17} achieved state-of-the-art performance. While also producing strong results, \cite{DBLP:journals/corr/DongYLMG17} took a convolutional neural network approach and implemented UNet \citep{DBLP:journals/corr/RonnebergerFB15}, a 3D, fully convolutional neural network. With these approaches performing well, 3D convolutional neural networks stand as the state-of-the-art approach to brain tumor segmentation with dice scores over 0.90. 

\subsection{Flaws and Issues: Deep Learning}

Even though the Random Forest and UNet yielded high dice scores, there are significant flaws with these approaches. Automated techniques not using convolutional neural networks typically have to incorporate hand-designed features. Due to the complexity of brain tumors and MRI imaging, it is difficult to select the correct features to hand-design, resulting in an excessive number of computations --- in other words, running out of GPU memory and increasing training time. In fact, learning features from the data rather than hand-crafting features has produced higher accuracies \citep{7153637}.

By using a convolutional neural network, the number of features is lowered by using dimensionality reduction, so training becomes much faster \citep{DBLP:journals/corr/HavaeiDWBCBPJL15}. However, certain CNN architectures still have their drawbacks. The UNet architecture takes 2D inputs, but networks with 2D inputs have been outperformed by networks with entire volume input as 2D MRI slices fail to capture all the scan information \citep{DBLP:journals/corr/KamnitsasLNSKMR16}. Networks similar to AlexNet \citep{NIPS2012_4824} have been successful at classifying images and segmenting brain tumors, as seen in the work of \cite{pereira2018adaptive}. Although the network produced a whole tumor dice score of 0.866, the method includes extensive preprocessing which drastically increases when dealing with large amounts of data \citep{DBLP:journals/corr/abs-1712-09093}. This problem will only grow in scale as more brain cancer datasets are published in the future. In order to achieve strong results, networks should be computationally efficient while requiring minimal preprocessing; take an MRI volume as an input; and have methods in place to counteract class imbalance.

\subsection{Flaws and Issues: MRI Imaging}
Not only are there issues with the current approaches to automated brain tumor segmentation with deep learning, but there are several flaws with MRI imaging. Although MRI imaging is able to show small details, it doesn't clearly differentiate healthy tissue from cancerous tissues in the brain. When a malignant tumor is in its early stages, the signs can be very subtle. If healthy tissue is mistaken for cancerous tissue, then the doctor may remove healthy brain tissue. If cancerous tissue is thought to be healthy, then the tumor will be able to spread should part of the tumor be left untouched. Despite the complexity of these challenges, careful neural network design can help address the flaws.

\section{V-Net}

With much success coming from a CNN approach, I decided to head in this direction. V-Net, proposed by Fausto Milletari et al., has produced a 0.869 dice score segmenting prostate images and the network takes an entire MRI volume as input which suits the task very well \citep{DBLP:journals/corr/MilletariNA16}. In this paper, I apply such a network to segment brain tumors in MRI scans. 

\subsection{Network Overview}

The V-Net network serves as a high-performing approach for volumetric image segmentation, and has produced results within the top ten networks in the PROMISE prostate segmentation competition \citep{LITJENS2014359}. The network takes an entire volume as its input, and --- as mentioned before --- this approach has outperformed inputting 2D slices of the MRI volume to the network. The V-Net network is split up into two ``sides,'' which pass their results to the other side. The network ultimately produces a segmentation of the brain tumor \citep{DBLP:journals/corr/MilletariNA16}. 

\subsection{Fit for the Problem}

Past convolutional neural network approaches split up volumes into 2D slices, and then performed patch-wise image classification. However, such methods solely consider local context and fail to take into account the entire image, making the models vulnerable to low performance. Additionally, these networks can have high inference times \citep{DBLP:journals/corr/KamnitsasLNSKMR16}. The dice loss layer addresses bias issues caused by class imbalance. Predictions are biased towards the background since the MRI scan mostly consists of healthy brain tissue, and the tumor (cancerous tissue) makes up a small portion of the scan. Therefore, there are more pixels that belong to the background (healthy tissue) class, so parts of the tumor may be classified as healthy tissue. As previously mentioned, the V-Net architecture produced strong results when tested on the PROMISE dataset \citep{LITJENS2014359} while demonstrating quick and efficient training. With this in mind, I decided to implement V-Net over many other previously existing network architectures.

\subsection{V-Net Architecture}

The V-Net architecture is a 3D, fully convolutional network which features a dice loss layer and convolutional operations over max pooling. The architecture can be seen in Figure 2 below. 

\begin{figure}[ht]
\centering
\includegraphics[scale = 0.08]{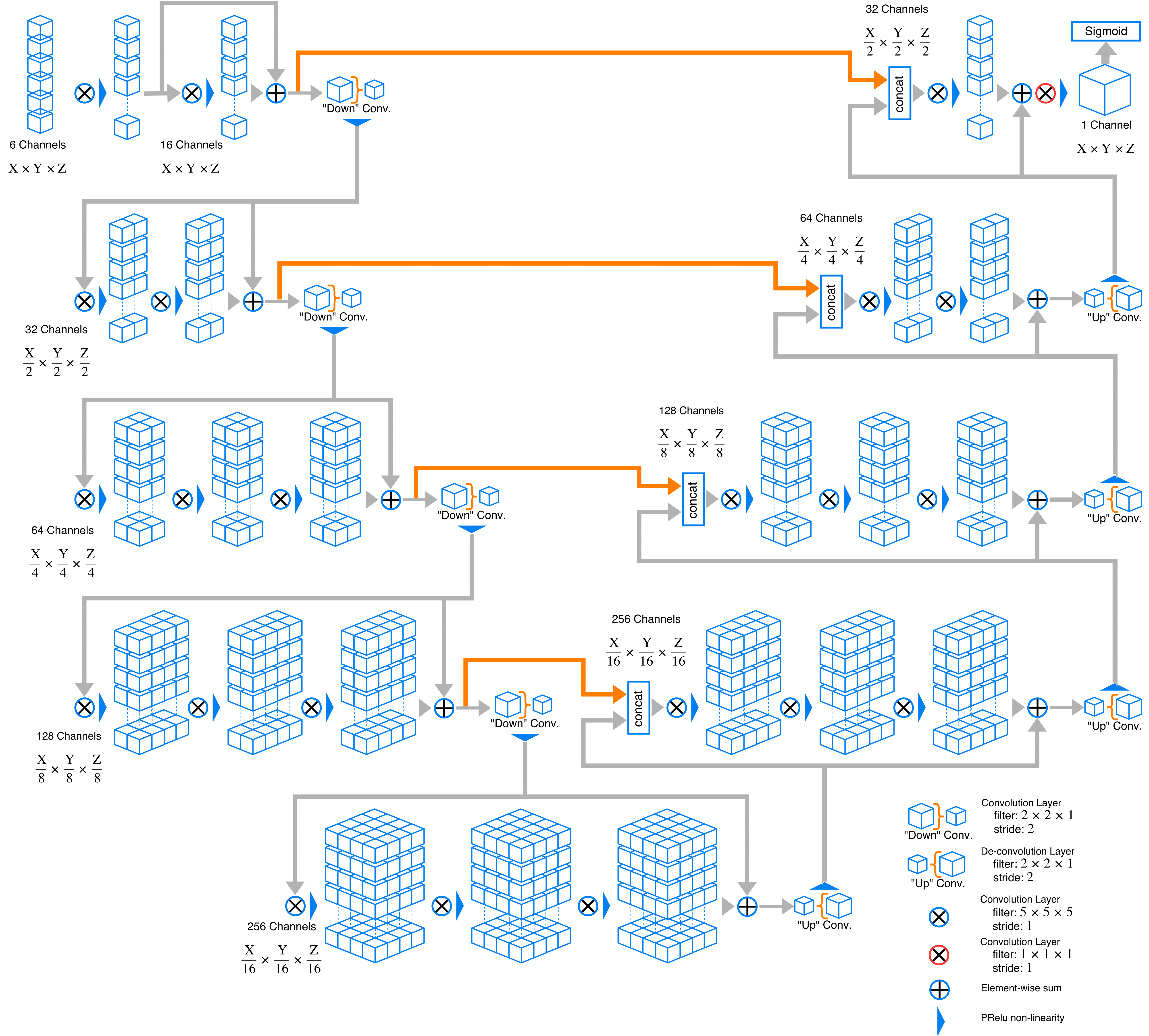}
\\
{\footnotesize \textbf{Fig. 2.} Schematic of the V-Net architecture \citep{DBLP:journals/corr/MilletariNA16}.\par}
\end{figure}
\FloatBarrier

The resolution is lowered as the data proceeds through all the steps in the left side of the network. Padding is applied to each convolution performed. On the right side of the network, the signal is decompressed until the signal reaches its original size; the other side is a compression path. Different stages that work at different resolutions make up the left side of the network. The stages consist of convolutional layers which learn a residual function. The output of each layer is passed to the next convolutional layer. This architecture enables convergence more quickly than networks that do not learn residual functions. The innermost layer of the network captures the whole input volume as the features are computed from a larger size than the size of the tumor \citep{DBLP:journals/corr/MilletariNA16}.

\begin{table}[ht]
\begin{tabular}{|l|l|l|l|l|l|}
\hline
Layer        & Input Size & Receptive Field & Layer         & Input Size & Receptive Field \\ \hline
Left Stage 1 & 128        & $5\times 5\times 5$       & Right Stage 4 & 16         & $476\times 476\times 476$ \\ \hline
Left Stage 2 & 64         & $22\times 22\times 22$    & Right Stage 3 & 32         & $528\times 528\times 528$ \\ \hline
Left Stage 3 & 32         & $72\times 72\times 72$    & Right Stage 2 & 64         & $546\times 546\times 546$ \\ \hline
Left Stage 4 & 16         & $172\times 172\times 172$ & Right Stage 1 & 128        & $551\times 551\times 551$ \\ \hline
Left Stage 5 & 8          & $372\times 372\times 372$ & Output        & 128        & $551\times 551\times 551$ \\ \hline
\end{tabular}
\caption{Receptive Field of the convolutional layers \citep{DBLP:journals/corr/MilletariNA16}.}
\end{table}
\FloatBarrier

Each convolution in the stages of the left side of the network uses kernels of size 5 x 5 x 5 voxels. Data flows through the different stages, and its resolution is reduced through convolution with $2\times 2\times 2$ voxel wide kernels. Features are extracted by only looking at $2\times 2\times 2$ patches that do not overlap, so the size of the feature maps are halved. Convolutional operations are used over max pooling after other works do not favor max pooling \citep{DBLP:journals/corr/abs-1711-01468, DBLP:journals/corr/FidonLGEKOV17aa}. Convolutional operations are used to double the number of feature maps as the resolution reduces, and PReLu non-linearities are applied all across the network \citep{DBLP:journals/corr/MilletariNA16}. 

By using convolutional operations over pooling operations, the memory footprint is lowered significantly since no switches mapping the pooling layers' output back to the inputs are needed for performing backpropagation. Through the use of downsampling, the signal size can be reduced while increasing the receptive field of the features in each layer of the network. The number of features is double the previous layer in each stage on the left side of the network. On the other side, the spatial support of the low resolution maps is expanded while also extracting features. By expanding the spatial support, a two-channel output segmentation can be completed. The final convolutional layer computes two feature maps with $1\times 1\times 1$ kernel size and produces outputs with the same size as the input volume so that they can be converted into probabilistic segmentations by applying the softmax function voxelwise. De-convolution operations are performed after each stage on the right half of the network which increases the size of the inputs. Next, there are three convolutional layers with half the number of kernels of the previous layer, and just like the left side of the network, residual functions are learned \citep{DBLP:journals/corr/MilletariNA16}.

The features extracted in the early stages of the left section of the network are forwarded to the right part to collect all the minute details that would have been lost in the compression path. By doing so, the network has a lower convergence time \citep{DBLP:journals/corr/MilletariNA16}.  

\subsection{Dice Loss Layer}
As previously discussed, class imbalance is an unavoidable problem since brain tumors are small relative to the rest of the MRI scan; there is more background than foreground. This imbalance can cause dangerous errors in the segmentation. Some approaches to the issue include reweighting the foreground to reduce the class imbalance, which requires a weight to be assigned to the different classes. However, a dice layer performs even better \citep{DBLP:journals/corr/MilletariNA16} \citep{DBLP:journals/corr/abs-1709-00382}.

\section{Data}

The MICCAI BraTS Challenge is an annual competition for medical experts and engineers, where participants are tasked with using machine learning and computer vision to segment brain tumors \citep{pmid25494501}\citep{pmid28872634}. Data used for my research was collected from the previous competitions. The data includes T1, T1c, T2, Flair, and ground truth images for training and testing which were reviewed by board-certified neuroscientists. Each of these scans have varying image contrasts which allow different regions of the brain to be viewed with more detail. These MRI scans feature both low and high grade glioblastoma (an aggressive cancer): the ground truth images are the real images of the tumor within the brain. In the 2017 competition, there were around 250 patients, so relative to other medical image datasets like the National Lung Cancer Screening Trial which contains 53,454 patients, the data size is limited \citep{pmid21045183}. Besides BraTS, public datasets for gliomas are unavailable,  and private datasets can be very difficult to access. This competition therefore serves as the sole source of experimentation and advancement in applying machine learning to the segmentation of gliomas.

\begin{figure}[h]
\centering
\includegraphics[scale = 0.5]{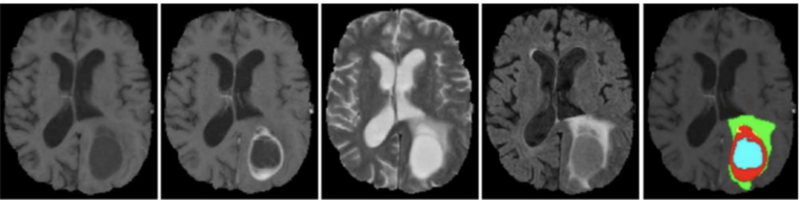}
\\
{\footnotesize \textbf{Fig. 1.} T1, T1C, T2, FLAIR, and ground truth images \citep{medium}.\par}
\end{figure}

\section{Preprocessing}

Preprocessing the data ensures uniformity, which helps improve network performance. The MRI images were Meta-Image files (.mha), and I converted them into NIfTI format to enable compatibility with Python medical imaging libraries. All scans were normalized to increase the consistency between the images. Next, I resized volumes to a uniform size to avoid any issues regarding a disparity in size between various volumes. Then, I applied a Gaussian noise mean filter to remove noise from the data before its input to the network. Noise occurs as a result of errors in the image acquisition process and produces an image with pixel values different from their true intensities, which can cause serious inaccuracies in the tumor location. The probability density function \textit{P} of a random variable x is given by:

$$
P(x) = \frac{1}{{\sigma \sqrt {2\pi } }}e^{{{ - \left( {x - \mu } \right)^2 } \mathord{\left/ {\vphantom {{ - \left( {x - \mu } \right)^2 } {2\sigma ^2 }}} \right. \kern-\nulldelimiterspace} {2\sigma ^2 }}}
$$
where $\mu$ is the mean and $\sigma$ is the standard deviation. Gaussian noise filtering is an effective method to reduce noise and boost performance.

\section{Training and Evaluation}

\subsection{Training}
Due to the intensive computations and significantly sized dataset needed to train the network, I used a NVIDIA Tesla K80 GPU with 12GB memory and 61GB of RAM. The network was trained in the cloud through an AWS Deep Learning AMI.

\subsection{S{\o}rensen-Dice Coefficient}

The S{\o}rensen-Dice Coefficient stands as the standard evaluation metric for medical image segmentation, prompting its use to evaluate the network performance. The S{\o}rensen-Dice Coefficient, D, can be defined as:

$$
D = \frac{2|X \cap Y|}{|X| + |Y|}
$$

Given two sets, X and Y, where $|X|$ and $|Y|$  are their respective cardinalities, the dice coefficient can be calculated.

\section{Results and Analysis}

\subsection{Results}

After two hundred epochs, the network was evaluated on a set of testing data which it had not seen before, where it received a whole tumor dice score of 0.89. In low training and inference times, and without major tweaking and fine-tuning of hyperparameters, the network was able to achieve state-of-the-art results. Given such characteristics, V-Net is well suited for clinical use. Clearly, V-Net has significant advantages over other networks for segmenting brain tumors in MRI scans.

\begin{figure}[h]
\centering
\includegraphics[scale = 0.2]{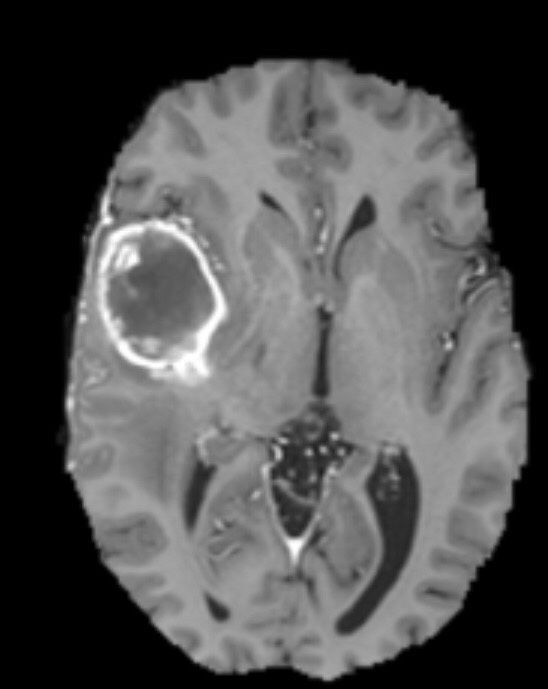}\includegraphics[scale = 0.2]{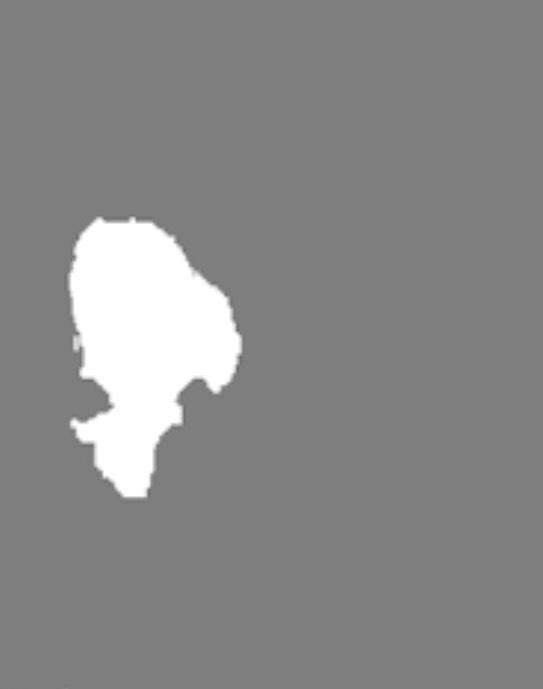}
\\
{\footnotesize \textbf{Fig. 3.} Results.\par}
\end{figure}

\subsection{Discussion of Past Results}

V-Net is a state-of-the-art architecture for brain tumor segmentation as it achieved a very high dice score that stands near the top of brain tumor segmentation performance.  While the results are not the highest performing of all methods, it outperforms most approaches (as seen in the table below).

\begin{table}[ht]
\centering 
\begin{tabular}{c c} 
Network & Whole Tumor Dice Score \\ [0.5ex] 
\hline 
\cite{DBLP:conf/miccai/JessonA17} & 0.86 \\ 
\cite{DBLP:journals/corr/KamnitsasLNSKMR16} & 0.901  \\
\cite{DBLP:journals/corr/ZhaoWSLZF17} & 0.87  \\
\cite{DBLP:journals/corr/abs-1709-00382} & 0.9050  \\
\cite{DBLP:journals/corr/abs-1708-00377} & 0.87  \\ [1ex] 
\hline 
\end{tabular}
\caption{Reults of other high performing networks.}
\label{table:nonlin} 
\end{table}

\bigskip
Most of the high-performing networks are some form of a convolutional neural network, but not all take an entire volume as input. \cite{DBLP:journals/corr/ZhaoWSLZF17} trained three different networks on 2D MRI slices of the axial, coronal, and sagittal views. With several different networks, training time is likely to be long. Using slices may also worsen the segmentation, since the number of pixels varies between slices. \cite{DBLP:journals/corr/abs-1709-00382} used dice loss which greatly improved performance and handled issues with an imbalance in training data. They built a cascade of fully convolutional neural networks which broke down the problem into three different binary segmentation tasks, and it placed 2nd in the MICCAI 2017 BraTS Challenge. A deep convolutional neural network was built by \cite{DBLP:journals/corr/abs-1708-00377} and was still able to achieve impressive results with a 0.87 dice score. While producing the lowest dice score out of these works, Jesson and Arbel's approach included a 3D Fully Connected Network (FCN) with reweighting to address class imbalance. With more specific modifications geared towards brain tumor segmentation, the network's performance can certainly be improved. To achieve a dice score of 0.901, \cite{DBLP:journals/corr/KamnitsasLNSKMR16} designed an 11-layer 3D CNN with a 3D fully connected Conditional Random Field (CRF) to remove false positives. The CRF proved to be effective at removing false positives and can certainly be included in networks in the future.

\subsection{Areas for Improvement}

Taking into consideration other approaches and the results, the performance of the V-Net can be further improved. I performed fairly little preprocessing on the data, so doing further preprocessing can help improve the network's performance. Something as simple as rotating the scans could boost results. As previously mentioned, automated segmentation software experiences low performance in a real-world setting due to the bias towards its training data. To combat this large problem, more brain tumor segmentation datasets must be constructed with diverse data, but that would likely take an extensive amount of time, money, and resources. To address low performance in practice, I later propose applying an existing technique for better results. 

\section{Conclusion and Future Work}

\subsection{Conclusion}

In the study, I applied a V-Net to segment brain tumors in MRI scans. The network used an entire volume as its input in order to decrease the training time, while preventing the network from putting too much emphasis on local context. As pointed out by other researchers, fully convolutional neural networks also seem to be the highest-performing for brain tumor segmentation and should be focused on by other researchers \citep{DBLP:journals/corr/KamnitsasLNSKMR16, DBLP:journals/corr/MilletariNA16, DBLP:journals/corr/DongYLMG17}. When segmenting a brain tumor, the region of interest is small relative to the size of the entire MRI scan. This causes issues in networks as too much emphasis is put on a small region, and the predictions are biased towards the background. To solve this issue, some design a loss function to reweight the foreground so that the prediction is not so biased towards the background. However, using dice loss over reweighting the foreground yields better results \citep{DBLP:journals/corr/MilletariNA16}. Fairly little preprocessing was done on the MRI scans, but extensive preprocessing will boost performance significantly despite taking longer. For instance, image enhancement alters the images so that they are more suitable for training. Although great results have been achieved by various researchers, there are still many issues with automated segmentation in medical practice.

\subsection{Future Work}

After an automated segmentation is completed, it is necessary to detect any errors in the segmentation since a significant error in the segmentation can be fatal for a patient. Since there is not an abundance of brain tumor segmentation datasets, models can easily overfit to the data it was trained on, causing many poor segmentations. However, without any ground truth, detecting errors is difficult. An RCA classifier can be built to address the problem \citep{DBLP:journals/corr/ValindriaLBKARR17}. It is trained on a network's predicted segmentations and ground truth images, with a dice score as the output. The RCA classifier was built with a CNN, Random Forest, and with atlas-based label propagation. The atlas-based label propagation easily outperformed the two other methods. When used in practice, the MRI scans can be taken and automatically segment the brain tumor, then the dice score can be predicted with the RCA classifier. This will let doctors know if they need to take a look at the segmentation and make manual adjustments to the automated segmentation. 

\begin{figure}[ht]
\centering
\includegraphics[scale = 0.5]{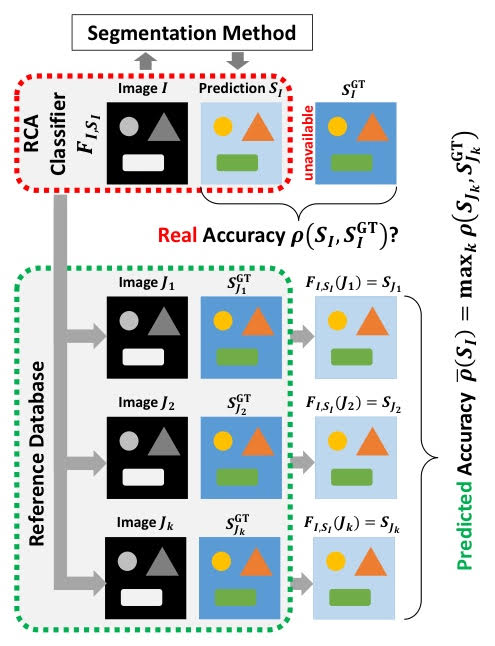}
\\
{\footnotesize \textbf{Fig. 4.} Diagram of the RCA architecture \citep{DBLP:journals/corr/ValindriaLBKARR17}.\par}
\end{figure}
\FloatBarrier

By combining an RCA classifier with a V-Net trained on MRI images of brain tumors into one piece of software, segmentations through deep learning can be more accurate and faster than ever, with the ability to save the lives of those living with brain cancer.

\newpage

\vskip 0.2in
\bibliography{sample}
\nocite{*}

\end{document}